\documentclass[11pt]{article}
\usepackage{coling2018}
\usepackage{times}
\usepackage{url}
\usepackage{latexsym}
\usepackage{color}
\usepackage{amsmath}
\usepackage{amssymb}
\usepackage{tikz}
\usepackage{subcaption}
\usepackage{wrapfig}

\newcounter{tbsnr}
\newenvironment{tbs}
{\addtocounter{tbsnr}{1}\par\bigskip\noindent\fbox{\thetbsnr}
\hspace*{\fill}\begin{minipage}{7cm}\tt}
{\end{minipage}\hspace*{\fill}\bigskip}

\newcommand{\cut}[1]{}

\title{Grounded Textual Entailment}

\author{Hoa Trong Vu$^{*+}$,  Claudio Greco$^\dagger$, Aliia Erofeeva$^{\dagger+}$, Somayeh Jafaritazehjan$^{*+}$\\
\textbf{Guido Linders$^{\dagger+}$, Marc Tanti$^*$,  Alberto  Testoni$^\dagger$,
 Raffaella Bernardi$^\dagger$, Albert Gatt$^*$}\\
 $^+$Erasmus Mundus European Program in Language and Communication
 Technology\\
$^*$University of Malta, $^\dagger$University of Trento \\
$^*${\tt name.surname@um.edu.mt},  $^\dagger${\tt name.surname@unitn.it}\\}

\date{}

\begin{document}
\maketitle

\begin{abstract}
Capturing semantic relations between sentences, such as entailment, is a long-standing challenge for computational semantics. Logic-based models analyse entailment in terms of possible worlds (interpretations, or situations) where a premise P entails a hypothesis H iff in all worlds where P is true, H is also true. Statistical models view this relationship probabilistically, addressing it in terms of whether a human would likely infer H from P. In this paper, we wish to bridge these two perspectives, by arguing for a visually-grounded version of the Textual Entailment task. Specifically, we ask whether models can perform better if, in addition to P and H, there is also an image (corresponding to the relevant ``world'' or ``situation''). We use a multimodal version of the SNLI dataset \cite{snli:emnlp2015} and we compare ``blind'' and visually-augmented models of textual entailment. We show that visual information is beneficial, but we also conduct an in-depth error analysis that reveals that current multimodal models are not performing ``grounding'' in an optimal fashion.
\end{abstract}

\section{Introduction}
\label{sec:introduction}
\blfootnote{Correspondence should be addressed to Raffaella Bernardi ({\tt raffaella.bernardi@unitn.it}) and Albert Gatt ({\tt albert.gatt@um.edu.mt}).}
\blfootnote{This work is licensed under a Creative Commons Attribution 4.0 International License. License
details: \\\url{http://creativecommons.org/licenses/by/4.0/}. The dataset, annotation and code is available from \\\url{https://github.com/claudiogreco/coling18-gte}.}

Evaluating the ability to infer information from a text is a crucial
 test of the capability of models to grasp meaning.  As a result,
the computational linguistics community has
invested huge efforts into developing textual entailment (TE) datasets.

After formal semanticists developed FraCas in the mid '90~\cite{coop:usin96},
an increase in statistical approaches to computational semantics 
gave rise to the need for suitable evaluation datasets.
Hence, Recognizing
Textual Entailment (RTE) shared tasks have been organized regularly~\cite{Sammons2012}.
Recent work on compositional distributional models has motivated the 
development of the SICK dataset
of sentence pairs in
entailment relations for evaluating such models~\cite{mare:sick14}.
Further advances with Neural Networks (NNs) have once more
motivated efforts to develop a large natural language inference
dataset, SNLI~\cite{snli:emnlp2015}, since NNs need to be trained on big data.  

However, meaning is
not something we obtain just from text and the ability to reason is
not unimodal either. The importance of enriching meaning
representations with other modalities has been advocated by
cognitive scientists, (e.g., ~\cite{Andrews2009,Barsalou2010}) and computational linguists
(e.g., ~\cite{Glavas2017}).
While efforts have been put into developing multimodal
datasets for the task of checking Semantic Text Similarity Text~\cite{Agirre2017}, 
we are not aware of any available datasets to tackle the problem of
Grounded Textual Entailment (GTE). Our paper is a first effort in this
direction.

\begin{figure}
\centering
\begin{subfigure}[b]{0.45\textwidth}
		\centering
         \includegraphics[scale=0.5]{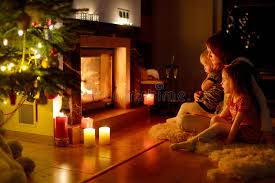}
          \caption{P: {\em A family in front of a chimney} and H: {\em A family
          trying to get warm}}
        \label{fig:ent}
    \end{subfigure}
\hfill
\begin{subfigure}[b]{0.45\textwidth}
		\centering
         \includegraphics[scale=0.5]{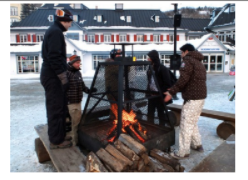}
        \caption{P: {\em People trying to get warm} and H: {\em People are outside on a
          chilly day}. TE: unrelated vs. GTE: related}
        \label{fig:neutral}
    \end{subfigure}
\caption{Premise, Hypothesis and Image examples}
\end{figure}

Textual Entailment is defined in terms of the likelihood of two
sentences (a premise P and an hypothesis H) to be in a certain
relation: P entails, contradicts or is unrelated to H. For instance,
the premise ``People trying to get warm in front of a chimney'' and
the hypothesis ``People trying to get warm at home'' are highly likely
to be in an entailment relation. Our question is 
whether having an image that
illustrates the event (e.g., Figure~\ref{fig:ent}) can help a model
to capture the relation. In order to answer this question, we augment the
largest available TE dataset with images, we enhance a state of the art
model of textual entailment to take images into account and we evaluate
it against the GTE task.

The inclusion of images can also alter relations which, based on text alone, would seem likely. For example, to a ``blind'' model the sentences of the sentence pair in Figure~\ref{fig:neutral} would seem to be unrelated, but
when the two sentences are viewed in the context of the image, they do become related.

A suitable GTE model therefore has to perform two sub-tasks: (a) it needs to 
ground its linguistic representations of P, H or both in non-linguistic (visual) data; (b) it needs to
reason about the possible relationship between P and H (modulo the visual information).

\section{Related Work}
\label{sec:related}
Grounding language through vision has recently become the focus of several tasks, including
Image Captioning (IC, e.g. ~\cite{Hodosh:etal:2013,xu:show15}) and Visual Question Answering
(VQA, eg.~\cite{mali:amul14,anto:vqa15}), and even more recently,
Visual Reasoning~\cite{girs:clev16,suhr-EtAl:2017:Short} and Visual
Dialog ~\cite{das:visdial2017}. Our focus is on Grounded Textual Entailment (GTE). While 
the literature on TE is rather vast,
GTE is still rather unexplored territory.

\paragraph{Textual Entailment}
Throughout the history of Computational Linguistics various datasets have
been built to evaluate Computational Semantics models on the TE task.
Usually they contain data divided into entailment, contradiction or
unknown classes. The ``unknown'' label has sometimes been replaced
with the ``unrelated'' or ``neutral'' label, capturing slightly
different types of phenomena. Interestingly, the ``entailment''
and ``contradiction'' classes also differ across datasets. In the
mid-'90s a group of formal semanticists developed FraCaS (Framework for
Computational Semantics).~\cite{coop:usin96}\footnote{A cleanly
  processable version of it has been made available only recently:
  \url{http://www-nlp.stanford.edu/~wcmac/downloads/fracas.xml}} The
dataset contains logical entailment problems in which a conclusion has
to be derived from one or more premises (but not necessarily all
premises are needed to verify the entailment). The entailments are
driven by logical properties of linguistic expressions, like the
monotonicity of quantifiers, or their conservativity property
etc. Hence, the set of premises entails the conclusion iff in all the
interpretations (worlds) in which the premises are true the conclusion
is also true; otherwise the conclusion contradicts the premises. In
2005, the PASCAL RTE (Recognizing Textual Entailment) challenge was
launched, to become a task organized annually.
In 2008, the RTE-4 committee made the task more
fine-grained by requiring the classification of the pairs as
``entailment'', ``contradiction'' and
``unknown''~\cite{giam:rte08}. The RTE datasets, unlike
FraCaS, contain real-life natural language sentences and the sort of
entailment problems which occur in corpora collected from the web.
Importantly, the sentence pair relations are annotated as entailment,
contradiction or neutral based on a likelihood condition: if
a human reading the premise would typically infer that the conclusion (called the
hypothesis) is most likely true (entailment), its negation is most
likely true (contradiction) or the conclusion can be either true or
false (neutral).

At SemEval 2014, in order to evaluate Compositional Distributional
Semantics Models focusing on the compositionality ability of those
models, the SICK dataset (Sentences Involving Compositional Knowledge)
was used in a shared entailment task ~\cite{mare:sick14}. Sentence pairs were obtained
through re-writing rules and annotated with the three RTE labels via a
crowdsourcing platform. Both in RTE and SICK the label assigned to the
sentence pairs captures the relation holding between the two
sentences. 

A different approach has
been used to build the much larger SNLI (Stanford Natural Language
Inference) dataset~\cite{snli:emnlp2015}: Premises are taken from a
dataset of images annotated with descriptive captions; the corresponding hypotheses
were produced through crowdsourcing, where for a given premise, annotators
provided a sentence which is true or not true with respect to a
possible image which the premise could describe. A consequence of this
choice is that the contradiction relation can be assigned to pairs
which are rather unrelated (``A person in a black wetsuit is surfing a
small wave'' and ``A woman is trying to sleep on her bed''),
differently from what happens in RTE and SICK.

Since the inception of RTE shared tasks, there has been an increasing emphasis on data-driven 
approaches which, given the hypothesis H and premise P, seek to classify the semantic relation (see~\cite{Sammons2012} 
for a review). More recently, neural approaches have come to dominate the scene, as shown by the 
recent RepEval 2017 task~\cite{Nangia2017}, where all submissions relied on bidirectional LSTM models, with or without pretrained embeddings. RTE also intersects with a number of related inference problems, including semantic text similarity and Question Answering, and some models have been proposed to address several such problems. In one popular approach, both P and H are encoded within the same embedding space, using a single RNN, with a decision made based on the encodings of the two sentences. This is the approach we adopt for our baseline LSTM in Section \ref{sec:models}, based on the model proposed by ~\newcite{snli:emnlp2015}, albeit with some modifications (see also \cite{Tan2016}). A second promising approach, based on which we adapt our state of the art model, relies on matching and aggregation~\cite{Wang2017}. Here, the decision concerning the relationship between P and H is based on an aggregate representation achieved after the two sentences are matched. Yet another area where neural approaches are being applied to sentence pairs in an entailment relationship is generation, where an RNN generates an entailed hypothesis (or a chain of such hypotheses) given an encoding of the premise~\cite{Kolesnyk2016,Starc2017}.

\paragraph{Vision and Textual Entailment}
In recent years, several models have been proposed to integrate the
language and vision modalities; usually the integration is
operationalized by element-wise
multiplication between linguistic and
visual vectors.
Though the interest in these modalities has spread in
an astonishing way thanks to various multimodal tasks proposed, including
 the IC, VQA, Visual Reasoning and Visual Dialogue tasks mentioned above,
very little work has been done on grounding entailment. Interestingly,
\newcite{youn:from14} has proposed the idea of considering images as
the ``possible worlds'' on which sentences find their
denotation. Hence, they released a ``visual denotation graph'' which
associates sentences with their denotation (sets of images). The idea 
has been further exploited by~\newcite{lai:lear17} and \newcite{han:visd17}.
\newcite{vend:orde16} look at hypernymy, textual entailment and image
captioning as special cases of a single visual-semantic hierarchy over
words, sentences and images, and they claim that modelling the partial
order structure of this hierarchy in visual and linguistic semantic
spaces improves model performance on those three tasks.  

We share
with this work the idea that the image can be taken as a possible
world. However, we don't use sets of images to obtain the visual
denotation of text in order to check whether entailment is logically
valid/highly likely. Rather, we take the image to be the
world/situation in which the text finds its interpretation.
The only work that is close to ours is an unpublished student
report~\cite{sitz:mult16}, which however lacks the in-depth analysis presented
here.

\section{Annotated dataset of images and sentence pairs}
\label{sec:dataset}

We took as our starting point the Stanford Natural Language Inference (SNLI)
dataset~\cite{snli:emnlp2015}, the largest natural language inference
dataset available with sentence pairs labelled with entailment, contradiction and neutral
relations. We augmented this dataset with images. It has
been shown very recently that SNLI contains language bias, such that a
simple classifier can achieve high accuracy in predicting the three
classes just by having as input the hypothesis sentence. A subset of
the SNLI test set with `hard' cases, where such a simplistic classifier fails (hereafter SNLI$_{hard}$)
has
been released~\cite{guru:anno18}. Hence, in this paper we will report
our results on both the full dataset and the hard test set, but then zoom
in on SNLI$_{hard}$
to understand the models' behaviour. We
briefly introduce SNLI and the new test set and compare
them through our annotation of linguistic phenomena.

\subsection{Dataset construction}

\paragraph{SNLI and SNLI$_{hard}$ test set} The SNLI dataset~\cite{snli:emnlp2015} 
was
built through Amazon
Mechanical Turk. Workers were shown captions of photographs 
without the photo
and were asked to write a new caption that (a) is definitely  a true description of the photo (entailment); (b) might be
a true description of the photo (neutral); (c) is definitely a false
description of the photo (contradiction). Examples were provided
for each of the three cases.
The premises are captions which come mostly from
Flickr30K ~\cite{youn:from14}; only 4K captions are from
VisualGenome ~\cite{kris:visu17}. In total, the dataset contains
570,152 sentence pairs, balanced with respect to the three
labels. Around 10\% of these data have been validated (4 annotators
for each example plus the label assigned through the previous data
collection phase). The development and test datasets contain 10K
examples each. Moreover, each Image/Flickr caption occurs in only one of the three
sets, and all the examples in the development and test sets have been
validated.

\paragraph{V-SNLI and V-SNLI$_{hard}$ test set} Our grounded version of SNLI, V-SNLI,
has been built by matching each sentence pair in SNLI with the
corresponding image coming from the Flickr30K dataset; thus the V-SNLI
dataset is slightly smaller than the original, which also contains
captions from VisualGenome. V-SNLI consists of 565,286
pairs (187,969 neutral, 188,453 contradiction, and 188,864
entailment). Training, test, and development splits have been built
according to the splits in SNLI. The main statistics of
the splits of the dataset are reported in Table \ref{table:vsnli}
together with statistics for the visual counterpart of Hard SNLI, namely
V-SNLI$_{hard}$. By construction, V-SNLI contains datapoints such that
the premise is always true with respect to the image, whereas the hypothesis
can be either true (entailment or neutral cases) or false
(contradiction or neutral cases.)

\begin{table}[!ht]
	\small
    \centering   
    \begin{tabular}{|l|r|r|r|r|}
        \hline
        \textbf{Split} & \textbf{\# entailment} & \textbf{\# contradiction} & \textbf{\# neutral} & \textbf{\# total} \\ \hline
        V-SNLI train & 182,167 & 181,938 & 181,515 & 545,620            \\ \hline
        V-SNLI dev & 3,329 & 3,278 & 3,235 & 9,842              \\ \hline
        V-SNLI test & 3,368 & 3,237 & 3,219 & 9,824              \\ \hline
        V-SNLI$_{hard}$ test & 1,058 & 1,135 & 1,068 & 3,261\\\hline
    \end{tabular}
     \caption{Statistics of the V-SNLI dataset.}
    \label{table:vsnli}
\end{table}

\subsection{Dataset annotation}
For deeper analysis and comparison of the contents of SNLI and SNLI$_{hard}$,
we have annotated the SNLI dataset by both automatically detecting some surface linguistic cues and manually labelling less trivial phenomena.
Using an in-house annotation interface, we collected human judgments aiming to (a) filter out those cases for which the gold-standard annotation was considered to be wrong\footnote{An example of a wrong annotation is the pair \emph{A white greyhound dog wearing a muzzle runs around a track} and \emph{The dog is racing other dogs}, labelled as entailment in the SNLI test set.};
(b) connect the three ungrounded relations to various linguistic phenomena.
To achieve this, we annotated a random sample of the SNLI test set containing 527 sentence pairs (185 entailment, 171 contradiction, 171 neutral),
out of which 176 were from the hard test set (56 entailment, 62 contradiction, 58 neutral).

All the pairs were annotated by at least two annotators, as follows:
(a) We filtered out all the pairs which had a wrong gold label (see Table~\ref{tab:changedlabels} for details).
When our annotators did not agree whether a given relation holds for a specific pair, we appealed to the corresponding five judgments coming from the validation stage of the SNLI dataset to reach a consensus based on the majority of labels. (b) We considered as valid any linguistic tag assigned by at least one annotator.
Since the annotation for (a) is binary whereas for (b) it is
multi-class, we used Cohen's $\kappa$ for the former and also Scott's $\pi$ and Krippendorff's $\alpha$ for the latter as suggested by Passonneau~\shortcite{pass:inte06}.
The inter-annotator agreement for the relation type (a) was $\kappa=0.93$; for (b) linguistic tags it was $\pi=0.63$, $\alpha=0.61$, and $\kappa=0.64$\footnote{Inter-rater agreement was calculated using the NLTK implementation, \url{http://www.nltk.org}}.

\begin{table}
    \small
    \begin{center}
        \begin{tabular}{|l|ccc|ccc|}
        	\hline
            & \multicolumn{3}{|c}{Ungrounded} & \multicolumn{3}{|c|}{Grounded}\\\hline
            & entailment & contradiction & neutral & entailment & contradiction & neutral\\\hline
            SNLI & 7\% & 16\% & 2\% & 1\% & 1\% & 31\% \\
            SNLI$_{hard}$ & 6\% & 10\% & 1\% & \textless1\% & 1\% & 20\%\\\hline
        \end{tabular}
        \caption{Wrong gold-standard labels: Data for which the gold standard label was considered to be wrong (a) in the \emph{ungrounded} setting or (b) correct in the ungrounded setting but not in the \emph{grounded} one. We filter out the data in (a) and keep those in (b). 
        }\label{tab:changedlabels}
    \end{center}
\end{table}

\paragraph{Linguistic phenomena}
Following the error analysis approach described
in recent work~\cite{Nangia2017,Williams2018}, we compiled a new list
of linguistic features that can be of interest when contrasting SNLI
and SNLI$_{hard}$, as well as for evaluating RTE models. Some of these
were detected automatically, while others were assigned manually. Automatic tags included \textsc{Synonym} and \textsc{Antonym}, which were detected using WordNet~\cite{Miller1995}.
 \textsc{Quantifier}, \textsc{Pronoun}, \textsc{Diff Tense}, \textsc{Superlative} and \textsc{Bare NP} were identified using Penn treebank labels~\cite{Marcus1993}, while labels such as \textsc{Negation} were found with a straightforward keyword search.
The tag \textsc{Long} has been assigned to sentence pairs with a premise containing more than 30 tokens,
or a hypothesis with more than 16 tokens.
Details about the tags used in the manual annotation are presented in Table~\ref{tab:manualtags}.

\begin{table}
	\small
    \centering
    \begin{tabular}{p{0.2\linewidth}p{0.4\linewidth}p{0.4\linewidth}}
    \hline
        Tag & Description & Example \\\hline
        Paraphrase & Two-way entailment, i.e., H entails P and vice versa. & P: \emph{A middle eastern marketplace}, H: \emph{A middle eastern store}\\
        Generalisation & One-way entailment, i.e., H entails P but not necessarily vice versa. & P: \emph{A group of people on the beach with cameras}, H: \emph{People are on a beach}.\\
        Entity & P and H describe different entities (e.g., subject, object, location) or incompatible properties of entities (e.g., color). & P: \emph{A dog runs along the ocean surf}, H: \emph{A cat is running in the waves}. \\
        Verb & The sentences describe different, incompatible actions. & P: \emph{Military personnel are shopping}, H: \emph{People in the military are training}. \\
        Insertion & H contains details and facts not present in P (e.g., subjective judgments and emotions.) & P: \emph{Woman reading a book in a laundry room}, H: \emph{The book is old}.\\
        Unrelated & The sentences are completely unrelated. & P: \emph{A woman is applying lip makeup to another woman}, H: \emph{The man is ready to fight}.\\
        Quantifier & The sentences contain numbers or quantifiers (e.g., \emph{all, no, some, both, group}). & P: \emph{A group of people are taking a fun train ride}, H: \emph{People ride the train}.\\
        World knowledge & Commonsense assumptions are needed to understand the relation between sentences (e.g., if there are named entities). & P: \emph{A crowd gathered on either side of a Soap Box Derby}, H: \emph{The people are at the race}.\\
        Voice & The premise is an active/passive transformation of the hypothesis. & P: \emph{Kids being walked by an adult}, H: \emph{An adult is escorting some children}.\\
        Swap & The sentences' subject and object are swapped from P to H. & P: \emph{A woman walks in front of a giant clock}, H: \emph{The clock walks in front of the woman}.\\\hline
    \end{tabular}
    \caption{Tags used in manual annotation of a subset of the SNLI test set.}\label{tab:manualtags}
\end{table}

We examined the differences in the tags distributions between the SNLI and SNLI$_{hard}$ test sets (Table~\ref{tab:tagsdistrib}).
Interestingly, the hard sentence pairs from our random sample include proportionately more antonyms but fewer pronouns, as well as examples with different verb tenses in the premise and hypothesis, compared to the full test set.
Furthermore, SNLI$_{hard}$ contains a significantly larger proportion of gold-standard labels which become wrong when the image is factored in ($\chi^2$-test with $\alpha=0.05$).

\begin{table}[tbp]
    \small
    \centering
    \begin{tabular}{|lrrrr|lrrrr|}
    \hline
        \multicolumn{1}{|c}{} & \multicolumn{2}{c}{SNLI} & \multicolumn{2}{c|}{SNLI$_{hard}$} & \multicolumn{1}{c }{} & \multicolumn{2}{c}{SNLI} & \multicolumn{2}{c|}{SNLI$_{hard}$} \\
        \multicolumn{1}{|l}{Manual tags} & \multicolumn{1}{c}{Freq} & \multicolumn{1}{c}{\%} & \multicolumn{1}{c}{Freq} & \multicolumn{1}{c}{\%} & \multicolumn{1}{|l}{Automatic tags} & \multicolumn{1}{c}{Freq} & \multicolumn{1}{c}{\%} & \multicolumn{1}{c}{Freq} & \multicolumn{1}{c|}{\%} \\\hline
        Insertion & 167 & 32 & 57 & 32 & \textsc{Diff Tense} & 7431 & 76 & 2384 & $\downarrow$74 \\
        Generalisation & 163 & 31 & 46 & 26 & \textsc{Quantifier} & 3779 & 39 & 1244 & 38 \\
        Entity & 107 & 20 & 37 & 21 & \textsc{Pronoun} & 3203 & 33 & 979 & $\downarrow$30\\
        Verb & 101 & 19 & 31 & 18 & \textsc{Synonym} & 1798 & 18 & 605 & 19 \\
        World knowledge & 93 & 18 & 34 & 19 & \textsc{Antonym} & 882 & 9 & 327 & $\uparrow$10\\
        Quantifier & 91 & 17 & 23 & 13 & \textsc{Superlative} & 304 & 3 & 106 & 3 \\
        Paraphrase & 7 & 1 & 4 & 2 & \textsc{Long} & 303 & 3 & 109 & 3 \\
        Unrelated & 6 & 1 & 2 & 1 & \textsc{Bare NP} & 281 & 3 & 107 & 3 \\
        Voice & 3 & 1 & 0 & 0 & \textsc{Negation} & 185 & 2 & 55 & 2 \\
        Swap & 1 & \textless1 & 1 & 1 & & \multicolumn{1}{l}{} & \multicolumn{1}{l}{} & \multicolumn{1}{l}{} & \multicolumn{1}{l|}{} \\\hline
    \end{tabular}
    \caption{Distribution of the automatic and manually assigned tags in the SNLI and SNLI$_{hard}$ test sets.
    Automatic tags are detected in the whole test set, manual ones are assigned to its random subset.
    Arrows $\uparrow\downarrow$ signify a statistically significant difference in tag proportions between the datasets (Pearson's $\chi^2$-test).}
    \label{tab:tagsdistrib}
\end{table}

\section{Models}
\label{sec:models}
 
In this section, we describe a variety of models that were compared on both V-SNLI and V-SNLI$_{hard}$, ranging from baseline models based on~\newcite{snli:emnlp2015} to a state of the art model by~\newcite{Wang2017}. We compare the original `blind' version of a model with a visually-augmented counterpart.
In what follows, we use {\em P} and {\em H} to refer to a premise and hypothesis, respectively.

\paragraph{LSTM baseline (Blind)} This model exploits a Recurrent Neural Network with Long Short-Term Memory units ~\cite{hochreiter1997:lstm} to encode both P and H in 512D vectors. The two vectors are then concatenated in a stack of three 512D layers having a ReLU activation function \cite{nair2010:relu}, with a final softmax layer to classify the relation between the two sentences as entailment, contradiction or neutral. The model is inspired by the LSTM baseline proposed by~\newcite{snli:emnlp2015}\footnote{There are some differences between our baseline and the LSTM baseline used in~\cite{snli:emnlp2015}. In particular, we used the Adam optimizer instead of the AdaDelta optimizer, the ReLU activation function instead of the tanh activation function, 512D instead of 100D for the output dimension of LSTM units, and dropout in all fully-connected layers instead of L2 regularization. Our settings outperformed the original ones in our experiments.}.
The model exploits the 300,000 most frequent pretrained GloVe embeddings \cite{Pennington2014} and improves them during the training process. To regularize the model, Dropout \cite{srivastava2014:dropout} is applied to the inputs and outputs of the recurrent layers and to the ReLU fully connected layers with a keeping probability of 0.5. The model is trained using the Adam
optimizer \cite{kingma2014:adam} with a learning rate of 0.001 until its accuracy on the
development set drops for three successive iterations.

\paragraph{V-LSTM baseline} The LSTM model described above is
augmented with a visual component following a standard Visual Question Answering baseline model~\cite{anto:vqa15}.  
Following initial representation of P and H in 512D vectors through an LSTM,
a fully-connected layer projects the L2-normalized 4096D image vector coming from the penultimate layer of a VGGnet16
Convolutional Neural Network~ \cite{simonyan2014:vggnet} to a reduced 512D vector.
A fully-connected layer with a ReLU activation function is also applied to P and H
to obtain two 512D vectors. 
The multimodal fusion between the text and the image is obtained
by performing an element-wise multiplication between the
vector of the text representation and the reduced vector of the
image. The multimodal fusion is performed between the image and both the premise and the hypothesis, resulting in two multimodal representations. The relation between them is captured as in the model described above. This model uses GloVe embeddings and the same optimization and procedure described above.

We have also adapted a state-of-the-art attention-based model for IC and VQA~\cite{Anderson2017up-down,teney2017:tips} to the GTE task. It obtains results comparable to the V-LSTM. This lack of improvement might be due to the need of
  further parameter tuning. We report the details of  our implementation
  and its results in the Supplementary Material.

\paragraph{BiMPM}
The Bilateral Multi-Perspective Matching (BiMPM) model~\cite{Wang2017} obtains state-of-the-art performance on the SNLI dataset, achieving a maximum
accuracy of 86.9\%, and going up to 88.8\% in an ensemble setup.
An initial embedding layer vectorises words in P and H using pretrained GLoVe embeddings \cite{Pennington2014}, and passing them to a context representation layer, which uses bidirectional LSTMs (BiLSTMs) to encode context vectors for each time-step. The core part of the model is the subsequent matching layer, where each contextual embedding or time-step of P is matched against all the embeddings of H, and vice versa.
The output of this layer is composed of two sequences of matching vectors, which constitute the input to another BiLSTM at the aggregation layer. The vectors from the last time-step of the BiLSTM are concatenated into a fixed-length vector, which is passed to the final prediction tier, a two-layer feed-forward network which classifies the relation between P and H via softmax.
Matching is performed via a cosine operation, which yields an $l$-dimensional vector, where $l$ is the number of perspectives. \newcite{Wang2017} experiment with four different matching strategies.
In their results, the best-performing version of the BiMPM model used all four matching strategies. We adopt this version of the model in what follows.

\paragraph{V-BiMPM model}
We enhanced BiMPM to account for the image, too. Our version of this
model is referred to as the {\em V-}BiMPM.
Here, the feature vector for an image is obtained from the layer before the fully-connected layer of a VGGnet-16. This results in a $7\times7\times512$ tensor, which we consider as 49 512-dimensional vectors.
The same matching operations are performed, except that matching occurs between P, H, and the image.
Since the textual and visual vectors have different dimensionality and belong to different spaces, we first map them to a mutual space using an affine transformation.
We match textual and image vectors using a cosine operation, as before. Full details of the model are reported in the Supplementary Materials for this paper.

\section{Experiments and Results}
\label{sec:experiment}

The models described in the previous sections were evaluated on both
(V-)SNLI and (V-)SNLI$_{hard}$. For the visually-augmented models, we
experimented with configurations where image vectors were combined
with both P and H (namely P+I and H+I), or only with H (P and
H+I). The best setting was invariably the one where only H was
grounded; hence, we focus on these results in what follows, comparing
them to ``blind'' models. In view of recent results suggesting that
biases in SNLI afford a high accuracy in the prediction task with only
the hypothesis sentence as input~\cite{guru:anno18}, we also include
results for the blind models without the premise (denoted with [H] in
what follows).

\begin{table}
\small
\begin{center}
\begin{tabular}{|l|c||cccc|}
\hline
  & LSTM [H] & LSTM & V-LSTM &   BiMPM &V-BiMPM \\\hline
Entailment &  72.65 &87.71 & 87.14 &  90.03 & 90.38  \\
Contradiction &  66.29 &79.7&  71.39  & 86.25 &  87.53  \\
Neutral & 66.36 & 76.79 &  68.06 &   82.79 & 82.91  \\\hline
Overall & 68.49 & 81.49 &  75.70 &   86.41 & \textbf{86.99} \\
\hline
\end{tabular}
\end{center}
\caption{Accuracies (\%) for V-SNLI. [H] indicates a baseline model encoding only the hypothesis.} \label{tab:snli}
\end{table}

\begin{table}
\small
\begin{center}
\begin{tabular}{|l|c||cccc|}
\hline
  & LSTM [H] & LSTM  & V-LSTM  & BiMPM &V-BiMPM  \\\hline
Entailment &  31.28 & 72.12&  69.09  &  80.43 & 81.38 \\
Contradiction & 25.29 & 60.79 &  46.34    & 77.62 & 76.12 \\
Neutral & 20.22 & 50.19 &  32.02 &   59.36 & 63.67\\\hline
Overall & 25.57 & 60.99 &   49.03 &   72.55 & \textbf{73.75} \\
\hline
\end{tabular}
\end{center}
\caption{Accuracies (\%) for V-SNLI$_{hard}$. [H] indicates a baseline model encoding only the hypothesis.} \label{tab:hard}
\end{table}

Table \ref{tab:snli} shows the results of the various models on the
full V-SNLI dataset. The same models are compared in Table
\ref{tab:hard} on V-SNLI$_{hard}$. First, note that the LSTM [H] model
evinces a drop in performance compared to LSTM (from 81.49\% to
68.49\%), though the drop is much greater on the unbiased
SNLI$_{hard}$ subset (from 60.99 to 25.57\%). This confirms the results reported by
\newcite{guru:anno18} and justifies our additional focus on this
subset of the data.

The effect of grounding in these models is less clear. The
LSTM baseline performs worse when it is visually augmented; this is the case of
V-SNLI and, even more drastically, V-SNLI$_{hard}$. It is also true
irrespective of the relationship type. On the other hand, the V-BiMPM
model improves marginally across the board, compared to BiMPM, on the
V-SNLI data. On the hard subset, the images appear to hurt performance
somewhat in the case of contradiction (from 77.62\% to 76.12\%), but
improve it by a substantial margin on neutral cases (from 59.36\% to
63.67\%).  The neutral case is the hardest for all models, with the
possible exception of LSTM [H] on the full dataset.

\cut{H+I A comparison of V-LSTM and V-BiMPM with their counterparts lacking the
premise (the [H+I]) cases shows that, as in the unimodal case,
performance of the grounded model drops when the premise is not
encoded, even in the presence of images. Once again, the drop is
larger on V-SNLI$_{hard}$ in both cases. Thus, the observation that
the premise matters in the TE task, and more so on a dataset that is
relatively free of annotation biases, holds true also of the grounded
case.} 

Overall, the results suggest that factoring in images either hinders performance (as in the case of the V-LSTM baseline), or helps only marginally (as in the case of V-BiMPM). In the latter case, we also observe instances where factoring in images hurts performance. In an effort to understand the results, we turn to a more detailed error analysis of the V-BiMPM model, first in relation to the dataset annotations, and then by zooming in somewhat closer on V-SNLI$_{hard}$.

\subsection{Error analysis by linguistic annotation label}
\begin{table}[!h]
\small
\centering
\begin{tabular}{|lrr|lrr|}
\hline
\multicolumn{1}{|c}{Manual tags} & \multicolumn{1}{c}{BiMPM} & \multicolumn{1}{c}{V-BiMPM} & \multicolumn{1}{|c}{Automatic tags} & \multicolumn{1}{c}{BiMPM} & \multicolumn{1}{c|}{V-BiMPM} \\\hline
Insertion & 58 & 63 & \textsc{Antonym} & 84 & 84 \\
Generalisation & 93 & 89 & \textsc{Bare NP} & 79 & 75 \\
Entity & 95 & $\downarrow$78 & \textsc{Quantifier} & 73 & 73 \\
Verb & 77 & 68 & \textsc{Diff Tense} & 72 & 73 \\
World knowledge & 79 & 71 & \textsc{Pronoun} & 69 & 70 \\
Quantifier & 78 & 70 & \textsc{Synonym} & 69 & 71 \\
Paraphrase & 75 & 75 & \textsc{Long} & 67 & 73 \\
Unrelated & 50 & 50 & \textsc{Superlative} & 64 & 63 \\
Swap & 0& 0& \textsc{Negation} & 51 & 56 \\
\hline
\end{tabular}
\caption{Accuracies obtained by BiMPM and V-BiMPM models on SNLI$_{hard}$, by annotation tags.  Arrows $\uparrow\downarrow$ signify a statistically significant difference in tag proportions between the datasets (Pearson's $\chi^2$-test).}\label{tab:tagseval}
\end{table}

In Table \ref{tab:tagseval}, accuracies for the blind and grounded
version of BiMPM are broken down by the labels given to the sentence
pairs in the annotated subset of SNLI described in Section
\ref{sec:dataset}. We only observe a significant difference  in the
{\em Entity} case, that is, where the referents in P and H are
inconsistent. Here, the blind model outperforms the grounded one, an
unexpected result, since one would assume a grounded model to be
better equipped to identify mismatched referents.  Hence, in the
following we aim to understand whether the models properly deal with
the grounding sub-task.

\subsection{Error analysis on grounding in the SNLI$_{hard}$}
We next turn to the ``hard'' subset of the data, where V-BiMPM showed some improvement 
over the blind case, but suffered on contradiction cases (Table \ref{tab:hard}). 
We analysed the 207 cases in SNLI$_{hard}$ where the
V-BiMPM made incorrect predictions compared to the blind model, that
is, where the image hurt performance. These were annotated
independently by two of the authors (raw inter-annotator agreement:
96\%) who (a) read the two sentences, P and H; (b) checked whether the
relation annotated in the dataset actually held or whether it was an
annotation error; (c) in those cases where it held, 
checked whether including the image actually resulted in a change in
the relation. 

Table \ref{tab:hard-mismatch} displays the proportions
of image mismatch and incorrect annotations. As the table suggests, in the cases where images hinder performance in
the V-BiMPM, it is usually because the image changes the relation
(thus, these are cases of image mismatch; see Section \ref{sec:introduction} for an example); this occurs in a large proportion of cases labelled
as neutral in the dataset.

\begin{table}
\small
\begin{center}
\begin{tabular}{|l|c|c|}
\hline
Relation & Image mismatch & Incorrect annotation \\
\hline
Entailment & 6.82 & 15.91 \\
Neutral & 44.58 & 1.20 \\
Contradiction & 3.80 & 22.78 \\
\hline
Overall & 24.76 & 12.62 \\
\hline
\end{tabular}
\caption{Cases where images hurt the V-BiMPM's performance: \% of
  cases in which including the image modifies the original SNLI relation (Image mismatch), and
 \% of cases in which the original SNLI relation is incorrectly annotated (Incorrect annotation).}\label{tab:hard-mismatch}
\end{center}
\end{table}

Inspired by the work in~\cite{miro:exam17}, we further explored the impact of visual grounding in both the V-LSTM
and V-BiMPM by comparing their performance on SNLI$_{hard}$, with the
same subset incorporating image ``foils''. Vectors for the images in the
V-SNLI test set were compared pairwise using cosine, and for each test
case in V-SNLI$_{hard}$, the actual image was replaced with the most
dissimilar image in the full test set.  The rationale is that, if
visual grounding is really helpful in recognising the semantic
relationship between P and H, we should observe a drop in performance
when the images are unrelated to the scenario described by the
sentences. The results are displayed in Table \ref{tab:wrongimages}, which also reproduces the original results on V-SNLI$_{hard}$ from Table \ref{tab:hard} for ease of reference.

As the results show, models are not hurt by the foil image, contrary to our expectations. 
V-BiMPM overall drops just by 0.67\% whereas V-LSTM
drop is somewhat higher (-2.11\%) showing it might be doing a better
job on the grounding sub-task.

 \begin{table}
\small
\begin{center}
\begin{tabular}{|l|cc|cc|}
\hline
& \multicolumn{2}{|c|}{V-LSTM} & \multicolumn{2}{c|}{V-BiMPM}\\ \hline
 & Original & Foil & Original & Foil  \\\hline
Entailment & 69.09& 65.03  & 81.38  & 80.81    \\
Contradiction & 46.34 & 30.92 &76.12 & 74.98  \\
Neutral & 32.02 & 31.46  & 63.67 & 63.39  \\
\hline
Overal & 49.03 &  46.92 (-2.11) & 73.75 & 73.08  (-0.67) \\
\hline
\end{tabular}
\caption{Accuracies of the visually-augmented models on
  V-SNLI$_{hard}$ with original or foil image.}\label{tab:wrongimages}
\end{center}
\end{table}

As a final check, we sought to isolate the grounding from the reasoning sub-task, focusing only on the former.
We compared the models
when grounding only the hypothesis [H+I], while leaving out the premise. Note that this test is
different from the evaluation of the model using only the hypothesis
[H]: Whereas in that case the input is not expected to provide any
useful information to perform the task, here it is. As we noted in
Section~\ref{sec:dataset}, by construction the premise is always true
with respect to the image while the hypothesis can be either true
(entailment or neutral cases) or false (contradiction or neutral
cases). A model that is grounding the text adequately would be expected to
confuse both entailment and contradiction cases with neutral ones; on the other hand, neutral cases
should be confused with entailments or contradictions. Confusing contradictions with entailments would be a sign that a
model is grounding inadequately, since it is not recognising that H is false with respect to the image.

As the left panel of Table \ref{tab:groundhyp} shows, V-BiMPM outperforms V-LSTM by a substantial margin, though the performance of both models drops substantially with this setup. The right panel in the table shows that neither model is free of implausible errors (confusing entailments and contradictions), though V-BiMPM makes substantially fewer of these. 

\begin{table}
\small
\begin{center}
\begin{tabular}{cc}
\begin{tabular}{|l|cc|}\hline
 & V-LSTM & V-BiMPM  \\ \hline
Entailment & 40.74& 51.89    \\ 
Contradiction & 30.22 & 40.7 \\
Neutral  & 22.47 & 32.02 \\\hline
Overall & 31.09 & 41.49 \\\hline
\end{tabular}

&

\begin{tabular}{|ll|cc|}\hline
GT class & Prediction & V-LSTM & V-BiMPM \\\hline
Contradiction & Contradiction & 343 & 462 \\
Contradiction & *Entailment &  442 & 327 \\
Contradiction & Neutral & 350 & 346  \\\hline
Entailment & Entailment & 431 & 549 \\
Entailment & *Contradiction & 254 & 166 \\
Entailment & Neutral & 373 & 343 \\\hline
Neutral & Neutral & 240 & 342 \\
Neutral & Contradiction &  377& 263 \\
Neutral & Entailment & 451 & 463 \\\hline
\end{tabular}
\end{tabular}
\caption{Confusion matrices for [H+I]. (*) marks implausible errors.}\label{tab:groundhyp}
\end{center}
\end{table}

\section{Conclusion}
\label{sec:conclusion}
This paper has investigated the potential of grounding the textual entailment task in visual data. We argued that a Grounded Textual Entailment model needs to perform two tasks: (a) the grounding itself, and (b) reasoning about the relation between the sentences, against the visual information. Our results suggest that a model based on matching and aggregation like the BiMPM model~\cite{Wang2017} can perform very well at the reasoning task, classifying entailment relations correctly much more frequently than a baseline V-LSTM. On the other hand, it is not clear that grounding is being performed adequately in this model.
It is primarily in the case of contradictions that the image seems to play a direct role in biasing the classification towards the right or wrong class, depending on whether the image is correct. 

In summary, two conclusions can be drawn from these results. First, in those cases where the inclusion of visual information results in a loss of accuracy, this is often due to the image resulting in a change in the original relation annotated in the dataset. A related observation is that using foil images results in a greater drop in performance on contradiction cases, possibly because in such cases, grounding serves to identify a mismatch between the hypothesis and the scene described by the premise, a situation which is rendered opaque by the introduction of foils. Second, in those cases where improvements are observed in the state of the art V-BiMPM, the precise role played by the image is not straightforward. Indeed, we find that this model still marginally outperforms the `blind', text-only model overall, when the images involved are foils rather than actual images.

We believe that further research on grounded TE is worthy of the NLP community's attention. While linking language with perception is currently a topical issue, there has been relatively little work on linking grounding directly with inference. By drawing closer to a joint solution to the  grounding and inference tasks, models will also be better able to address language understanding in the real world. 

The present paper presented a first step in this direction using a version of an existing TE dataset which was augmented with images that could be paired directly with the premises, since these were originally captions for those images. However, it is important to note that in this dataset premise-hypotheses pairs were not generated directly with reference to the images themselves. An important issue to consider in future work on GTE, besides the development of better models, is the development of datasets in which the role of perceptual information is controlled, ensuring that the data on which models are trained represents truly grounded inferences.

\section*{Acknowledgements}
 We kindly acknowledge the European Network on Integrating Vision and Language (iV\&L Net) ICT COST Action IC1307. Moreover, we thank the Erasmus Mundus European Program in Language and Communication Technology. Marc Tanti's work is partially funded by the Endeavour Scholarship Scheme (Malta),  part-financed by the European Union's European Social Fund (ESF). Finally, we gratefully acknowledge the support of NVIDIA Corporation with the donations to the University of Trento of the GPUs used in our research.

\section*{Appendix A: Bottom-up top-down attention (VQA)}
We adapted the Visual Question Answering model proposed
in~\cite{Anderson2017up-down,teney2017:tips} to the Grounded Textual
Entailment task. The model presents a more fine-grained attention
mechanism which allows to identify the most important regions
discovered in the image and to perform attention over each of
them.

The model uses a a Recurrent Neural Network with Long Short-Term Memory units to encode the premise P and hypothesis H in 512D vectors. A bottom-up attention mechanism exploits a Fast R-CNN \cite{girshick2015:fastrcnn} based on
a ResNet-101 convolutional neural network \cite{he2016:resnet} to
obtain region proposals corresponding to the 36 most informative
regions of the image. A top-down attention mechanism is used
between the premise (resp. hypothesis) and each of the L2-normalized 2048D
image vectors corresponding to the region proposals to obtain an
attention score for each of them. Then, a 2048D image vector encoding
the most interesting visual features for the premise (hypothesis) is
obtained as a sum of the 36 image vectors weighted by the
corresponding attention scores for the premise (hypothesis). A
fully-connected layer with a gated tanh activation function is applied
to the image vector of the most interesting visual features for the
premise and for the hypothesis to obtain a reduced 512D vector for
each of them. A fully-connected layer with a gated tanh activation
function is also applied to the premise and to the hypothesis in order
to obtain a reduced 512D vector for each of them. 

The multimodal
fusion between the premise (hypothesis) and the image vector of the
most interesting visual features for the premise (hypothesis) is
obtained by performing an element-wise multiplication between the
reduced vector of the premise (hypothesis) and the reduced vector of
the most interesting visual features for the premise
(hypothesis). After that, the model feeds the concatenation of the two
resulting multimodal representations to a stack of three 512D layers
having a gated tanh activation function, with a final softmax layer to
classify the relation between the two sentences as entailment,
contradiction or neutral. This model uses GloVe embeddings and the
same optimization tricks and procedure of the LSTM and V-LSTM models.

We report the accuracies of the VQA models against the various tests
reported in the paper. For ease of comparison we reproduce the full table from the main paper,
with the addition of the  VQA results.

\begin{table}[!ht]
	\begin{center}
		\begin{tabular}{|l|c||ccccc|}
			\hline
			& LSTM [H] & LSTM & V-LSTM &   VQA & BiMPM &V-BiMPM   \\\hline
			Entailment &  72.65 &87.71 & 87.14 & 86.1 & 90.03 & 90.38    \\
			Contradiction &  66.29 &79.7&  71.39   & 78.99& 86.25 &  87.53  \\
			Neutral & 66.36 & 76.79 &  68.06 &  73.56 & 82.79 & 82.91  \\\hline
			Overall & 68.49 & 81.49 &  75.70 &  79.65  & 86.41 & \textbf{86.99} \\
			\hline
		\end{tabular}
	\end{center}
	\caption{Accuracies (\%) for V-SNLI. [H] indicates a baseline model encoding only the hypothesis (Table \ref{tab:snli} in the paper).} \label{tab:snli-supp}
\end{table}

\begin{table}[!ht]
	\begin{center}
		\begin{tabular}{|l|c||ccccc|}
			\hline
			& LSTM [H] & LSTM  & V-LSTM  & VQA & BiMPM &V-BiMPM  \\\hline
			Entailment &  31.28 & 72.12&  69.09  & 67.39&   80.43 & 81.38 \\
			Contradiction & 25.29 & 60.79 &  46.34    & 59.03 & 77.62 & 76.12 \\
			Neutral & 20.22 & 50.19 &  32.02 &  42.13  & \textbf{59.36} & \textbf{63.67}\\\hline
			Overall & 25.57 & 60.99 &   49.03 &  56.21  & 72.55 & 73.75 \\
			\hline
		\end{tabular}
	\end{center}
	\caption{Accuracies (\%) for V-SNLI$_{hard}$. [H] indicates a baseline model encoding only the hypothesis (Table 6 in the paper).} \label{tab:hard-supp}
\end{table}

\begin{table}[!ht]
\begin{center}
\begin{tabular}{|l|cc|cc|cc|}
\hline
& \multicolumn{2}{|c|}{V-LSTM} & \multicolumn{2}{c|}{V-BiMPM} & \multicolumn{2}{c|}{VQA}\\ \hline
 & Original & Foil & Original & Foil   & Original & Foil \\\hline
Entailment & 69.09& 65.03  & 81.38  & 80.81    & 67.39& 60.4\\
Contradiction & 46.34 & 30.92 &76.12 & 74.98  & 59.03& 60.97\\
Neutral & 32.02 & 31.46  & 63.67 & 63.39   & 42.13& 42.79 \\
\hline
Overal & 49.03 &  46.92 (-2.11) & 73.75 & 73.08  (-0.03) & 56.21 &
                                                                   54.83 (-1.38) \\
\hline
\end{tabular}
\caption{Accuracies of the visually augmented models on V-SNLI$_{hard}$ containing the original or foil image (Table 9 in the paper).} \label{tab:wrongimages-supp}
\end{center}
\end{table}

\begin{table}[!ht]
\small
\begin{center}
\begin{tabular}{cc}
\begin{tabular}{|l|ccc|}\hline
 & V-LSTM & V-BiMPM  & VQA\\ \hline
Entailment & 40.74& 51.89   & 48.11 \\ 
Contradiction & 30.22 & 40.7 & 47.05\\
Neutral  & 22.47 & 32.02 & 31.37\\\hline
Overall & 31.09 & 41.49 & 42.26\\\hline
\end{tabular}
&


\begin{tabular}{|ll|ccc|}\hline
GT class & Prediction & V-LSTM & V-BiMPM & VQA\\\hline
Contradiction & Contradiction & 343 & 462 & 534\\
Contradiction & *Entailment &  442 & 327 & 286\\
Contradiction & Neutral & 350 & 346 & 315 \\\hline
Entailment & Entailment & 431 & 549 & 509\\
Entailment & *Contradiction & 254 & 166 & 188\\
Entailment & Neutral & 373 & 343 & 361\\\hline
Neutral & Neutral & 240 & 342 & 335\\
Neutral & Contradiction &  377& 263 & 272\\
Neutral & Entailment & 451 & 463 & 461\\\hline
\end{tabular}
\end{tabular}
\caption{Confusion matrices for [H+I]. (*) marks implausible errors (Table 10 in the paper).} \label{tab:groundhyp-supp}
\end{center}
\end{table}

%
%
\pagebreak
\section*{Appendix B: V-biMPM Model details}
\input{figures/vbimpm.tikz}

Here, we report some further details of our implementation of the V-BiMPM model described in Section~4 of the main paper, based on the work of 
\newcite{Wang2017}. Our model is displayed in Figure~\ref{fig:vbimpm}. 

The core part of the original BiMPM  is the matching layer. Given two $d$-dimensional vectors $v_{P}$ and $v_{H}$, each replicated $l$ times ($l$ is the number of `perspectives') and a trainable $l \times d$ weight matrix $W$, matching involves a cosine similarity computation that yields an $l$-dimensional matching vector $m$, whose elements are defined as follows:

\begin{equation}
m_{k} = \textit{cosine}\left(W_{k} \circ v_{P}, W_{k} \circ v_{H}\right)
\end{equation}

\noindent
The matching operations included are the following:

\begin{enumerate}
\item {\em full-matching}, where each forward or backward contextual embedding of the premise P (resp. the hypothesis H) is matched to the last time-step of H (resp. P); 
\item {\em max-pooling}, where each forward/backward contextual embedding of one sentence is compared to the embeddings of the other, retaining the maximum value for each dimension; 
\item {\em attentive matching}, where first, the pairwise cosine similarity between forward/backward embeddings of P and H is estimated, before calculating an attentive vector over the weighted sum of contextual embeddings for H and matching each forward/backward embedding of P against the attentive vector;
\item {\em max-attentive matching}, a version of attentive matching where the contextual embedding with the highest cosine is used as the attentive vector, instead of the weighted sum. 
\end{enumerate}

\noindent
The visually-augmented version of the original model, V-BiMPM, is displayed in Figure \ref{fig:vbimpm}. To perform multimodal matching, the visual and textual vectors are mapped to a mutual space using the following affine transformation:

\begin{equation}
v_{i} = \mathbf{W}_{t}f_{i}+b_{t}; f_{i} \in \mathbb{R}^{e}; \mathbf{W}_{t} \in \mathbb{R}^{e \times d}; b_{t}, v_{i} \in \mathbb{R}^{d}
\end{equation}

\noindent 
where $\mathbf{W}_{t}$, $b_{t}$,  $f_{i}$, and $v_{i}$ are the weight matrix, the bias, the input features and output features, respectively, and $t$ is any text (P or H). Given weight matrices $\mathbf{W} \in \mathbb{R}^{l \times d}$ for text and $\mathbf{U}^{l \times d}$ for images, we compute the matching vector $m$ between a textual vector $v_{t}$ and image vector $v_{i}$ as:

\begin{equation}
m_{k} = \textit{cosine}\left(W_{k} \circ v_{t}, U_{k} \circ v_{i}\right)
\end{equation}



\bibliographystyle{acl}
\bibliography{raffa}
\end{document}